\documentclass[runningheads]{llncs}

\usepackage{graphicx}
\usepackage{amssymb}
\usepackage{rotating}

\usepackage{float}
\usepackage{subfigure}

\usepackage{subfigure}
\usepackage{wrapfig}
\usepackage{booktabs}
\usepackage{multirow}
\usepackage{makecell}
\usepackage{enumitem}

\usepackage{bbding}
\usepackage{xcolor}
\usepackage{bm}
\usepackage{amsmath}

\usepackage[capitalise]{cleveref}

\begin{document}

\title{A Survey of Explainable Graph Neural Networks: Taxonomy and Evaluation Metrics}
\titlerunning{A Survey of Explainable GNNs: Taxonomy and Evaluation Metrics}

\author{Yiqiao Li \and Jianlong Zhou \and Sunny Verma \and Fang Chen}

\institute{University of Technology Sydney, Sydney, Australia\\
\email{yiqiao.li-1@student.uts.edu.au}\\
\email{\{jianlong.zhou, sunny.verma, Fang.Chen\}@uts.edu.au}
}

\maketitle             
\begin{abstract}

Graph neural networks (GNNs) have demonstrated a significant boost in prediction performance on the graph data. At the same time, the predictions made by these models are often hard to interpret. In that regard, many efforts have been made to explain the prediction mechanisms of these models from perspectives such as GNNExplainer, XGNN and PGExplainer. Although such works present systematic frameworks to interpret GNNs, a holistic review for explainable GNNs is unavailable. In this survey, we present a comprehensive review of explainability techniques developed for GNNs. We focus on explainable graph neural networks, categorize them based on the use of explainable methods. We further provide the common performance metrics for GNNs explanations and point out several future research directions.

\end{abstract}


\section{Introduction}\label{sec:intro}

A graph $\mathcal{G}$ can be viewed as a representation of certain relationship formed by a set of nodes $\mathcal{N}_i$ ($i=1, 2, \cdots, n$) and edges $\mathcal{E}_j$ ($j=1, 2, \cdots, m$). It is an ideal data structure that can be used to model a variety of real-world datasets (e.g., molecules). With the resurgence of deep learning, graph neural networks (GNNs) have been a powerful tool to model graph data and achieved impressive performance in a great deal of domains and applications, such as recommendation, chemistry, medical, and etc.~\cite{wu2020graph,goh2017smiles2vec,tjoa2020survey}. However, incorporating both graph structure and feature information together has lead to complex non-linear models, increasing the difficulties of understanding its working mechanism as well as its predictions. On the other hand, an explainable model is favored and even necessary, especially in practical scenarios (e.g., medical diagnosis), as explanations benefit users in multiple ways such as improving the model's fairness/security, and it also enhance trust in the model's recommendations. As a result, eXplainable GNNs (XGNNs) has achieved considerable research attention in the recent years and can be categorized into two categories: 1) making the eXplainable-AI (XAI) methods and using them directly to explain GNNs; 2) developing their strategies based on graph intrinsic structures and features and doing not involving the XAI approaches.



Although there is an increasing number of work that focus on the explainability of GNNs in recent years, there is few systematical discussion about them. We believe that analyzing these recent work of XGNNs in a comprehensive way would facilitate a better understanding of these methods, stimulate new ideas, and provide insight of developing new explainable methods. Therefore, we analyze and summarize the current methods of explainable methods for GNNs. In particular, we categorize them into two groups---XAI-based XGNNs in Section~\ref{sec:XAI-GNNs} and non-XAI-based XGNNs in Section~\ref{sec:XGNNs}. We then present the metrics that are used to measure the explainability of XGNNs in Section~\ref{sec:metric}. We discuss recurrent issues with XGNNs in Section~\ref{sec:dis}, and finally point out several future research directions in Section~\ref{sec:con}.

Our contributions can be summarized as:

\begin{itemize}
    
    \item We systematically analyze state-of-the-art methods of XGNNs and categorize them into two groups: \emph{XAI-based XGNNs}, which leverage the existing XAI approaches to explain GNNs; \emph{Non-XAI-based XGNNs}, which moves aways from current XAI methods while attempts to explain GNNs through taking advantage of the inherent structures and features of graphs. 
    
    \item We present the evaluation metrics for XGNNs, which can be used to measure the performance of XGNNs methods, as knowledge of evaluation metrics are necessary to educate the end-users/practitioners of XGNNs.  
    
    
    \item We discuss the recurrent problems in the filed of XGNNs along with possible solutions, and finally point out several potential research directions to further improve the explainability of GNNs.

\end{itemize}

\begin{figure}
\centering
\includegraphics[width=0.65\textwidth]{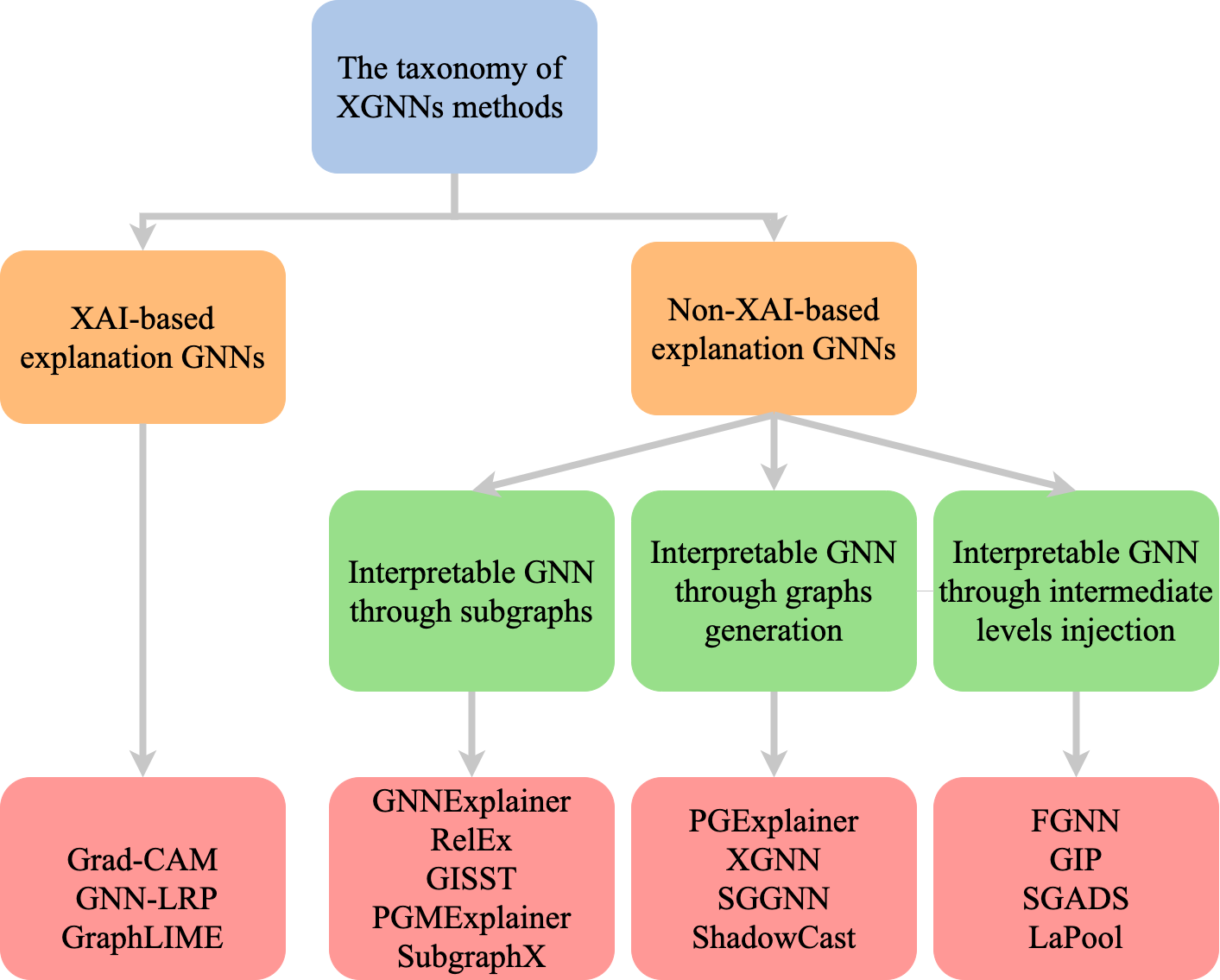}
\caption{The taxonomy of explainable graph neural networks.}
\label{fig:xggns}
\end{figure}

\section{XAI-based Explainable Graph Neural Networks}\label{sec:XAI-GNNs}

By analyzing the references of XGNNs, we made a binary classification of explainable GNNs’ methods, which can be divided into two categories: XAI-based and non-XAI-based. The taxonomy of XGNNs is shown in Figure~\ref{fig:xggns}. We begin by presenting a brief introduction of XAI and then present XGNNs, as it will aid understanding of XAI-based explainable techniques for XGNNs.

\subsection{Explainable Artificial Intelligence}\label{subsec:XAI}

Over the past years, XAI has been becoming a hot research topic and there is an increasing number of studies in this field. Several surveys have summarized the history, taxonomy, evaluation, challenges and opportunities about it, mainly focusing on the explanation of deep neural networks (DNNs)~\cite{adadi2018peeking}\cite{chakraborti2020emerging}\cite{das2020opportunities}\cite{srinivasanexplanation}\cite{guidotti2018survey}. 

\begin{figure}
    \centering
    \includegraphics[width=0.95\textwidth]{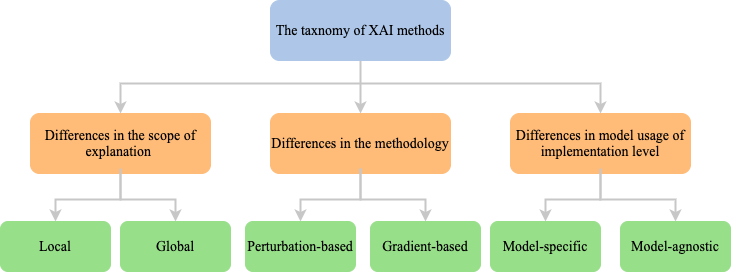}
    \caption{The taxonomy of XAI methods.}
    \label{fig:XAI taxnomy}
\end{figure}

XAI techniques can be classify according to three categories as discussed in~\cite{das2020opportunities}: (i) the difference of scope of interpretability, (ii) the difference of methodology, and (iii) the difference of usage of ML models (see Figure~\ref{fig:XAI taxnomy}). 

We can also divide XAI into \textit{model-specific} XAI and \textit{model-agnostic} XAI, based on the difference of usage of ML model. The model-specific XAI refers to any methods that focus on the explainability of a single or a group of specific AI models; while the model-agnostic XAI does not put any emphases on the underlying AI models.

Model-agnostic XAI can be used to assess most AI models and are often applied after the training, thus, they are usually treated as a post-hoc method. Model-agnostic XAI relies on analyzing pairs of input and output features and has no access to the specific inner workings of AI models (e.g., weights or structural information), otherwise it will be impossible to decouple it from the black-box models~\cite{stiglic2020interpretability}. After analyzing the characteristics of model-specific XAI and model-agnostic XAI, we can see that, model-specific XAI methods heavily rely on the specific parameters, while any changes in the architecture of models may result in significant changes in the interpretation method itself or the corresponding explainable algorithm. Thus, the model-specific XAI methods are not available to extend to explain GNN. However, some model-agnostic XAI methods can be extended to explain GNNs.

\subsection{Explaining Graph Neural Networks through XAI Methods }\label{subsec:XAI-GNNS}

Convolutional neural networks (CNNs) could be used in the graph-structured data by extending the convolution operation onto graphs and in general onto non-Euclidean spaces. The extension of CNNs to non-Euclidean spaces is regared as graph convolution neural networks (GCNNs). Thus, we can adapt the common explainability methods which are originally designed for CNNs, and extend them to GCNNs. And we found that a variety of methods of XAI can be easily extended to GNNs, such as LRP~\cite{bach2015pixel}, LIME~\cite{ribeiro2016should}, Grad-CAM~\cite{selvaraju2017grad}. These extensions are summarized in Table~\ref{tab:XAIandXGNNs}.

\begin{table}[!htpb]
\caption{The extensions of XAI to GNNs.}\label{tab:XAIandXGNNs}
\centering
\setlength{\tabcolsep}{18mm}{
\begin{tabular}{c  c }
\toprule

XAI Methods & XGNNs Methods\\
\midrule
LRP & GNN-LRP~\cite{sun2020explanation,schnake2020xai,cho2020interactionnet}  \\
Grad-CAM & Grad-CAM~\cite{pope2019explainability} \\
LIME & GraphLIME~\cite{huang2020graphlime} \\
 
\bottomrule
\end{tabular}
}

\end{table}

\textbf{Layer-wise Relevance Propagation (LRP)} assumes that the classifier can be decomposed into several computational layers, and propagates the DNNs output from the top layer to the input layer. At each layer, a propagation rule is applied ~\cite{bach2015pixel}. The contributions to the target output node are back-propagated to the input features form a map of features that contribute to that node. Therefore, LRP is useful to visualize the contributions of input features to models predictions, especially for kernel-based classifiers and multi-layer neural networks. 

Motivated by it, the researchers~\cite{sun2020explanation} used LRP in GNN to obtain insights into the black-box of GNN models. Schnake et al.~\cite{schnake2020xai} proposed GNN-LRP based higher-order Taylor expansions. GNN-LRP produces detailed explanations that subsume the complex nested interaction between the GNN model and the input graph. Furthermore, Cho et al.~\cite{cho2020interactionnet} conducted the post-hoc explanation on individual predictions with the use of LRP. The LRP calculates the relevance for every neuron by reversely propagating, through the network, from the predicted output to the input level, and the relevance represents the quantitative contribution of a given neuron to the prediction. What’s more, Baldassarre et al.~\cite{baldassarre2019explainability} also applied LRP to graph models. The LRP method computes the saliency maps by decomposing the output prediction into a combination of its inputs. 

\textbf{Local Interpretable Model-Agnostic Explanations (LIME)} is another popular method in XAI. LIME extracts individual prediction instances from the black-box model and generates a simpler and explainable model such as linear model to approximate the decision features of it. This simple model can then be interpreted and used to explain the original black-box predictions~\cite{ribeiro2016should}. Many other papers have improved and extended the LIME. Zhao et al.~\cite{zhao2020baylime} introduced BayLIME that incorporates LIME with Bayesian. Zafar et al.~\cite{zafar2019dlime} used the Jaccard similarity among multiple generated explanations and proposed a deterministic version of LIME. Furthermore, LIME has been widely applied in GNNs to explain GNN models. Huang et al.~\cite{huang2020graphlime} proposed GraphLIME, a local interpretable model explanation for graphs using the Hilbert-Schmidt Independence Criterion (HSIC) Lasso, which is a nonlinear feature selection method to achieve local explainability. Their frameworks are generic GNN-model explanation framework that learns a nonlinear interpretable model locally in the subgraph of the node being explained.

\textbf{Gradient-weighted Class Activation Mapping (Grad-CAM)} improves CAM by relaxing the architectural restriction that the penultimate layer must be convolutional~\cite{selvaraju2017grad}. It generates a coarse localization map to highlight the important regions in the input images by making use of the gradients of the target concept flowing into the final convolutional layer. Grad-CAM has been applied to a wide variety of convolutional neural network modelfamilies~\cite{selvaraju2017grad}. Pasa~\cite{pasa2019efficient} directly used it as a visualization tool for the convolutional neural network explanations. Vinogradova et al.~\cite{vinogradova2020towards} further extended the Grad-CAM and applied it locally to produce heatmaps showing the relevance of individual pixels in semantic segmentation. Grad-CAM can also be extended to GNN. Pope et al.~\cite{pope2019explainability} described the extension of CNN explainable methods to GCNNs. They introduced explainable method (Grad-CAM) for decisions made by GCNNs. Grad-CAM enables to generate heat-maps with respect to different layers of the network.




\section{Non-XAI-based Explainable Graph Neural Networks}\label{sec:XGNNs}

Most XAI-based methods for XGNNs do not require knowledge of the internal parameters of the GNN model and the XAI methods used to yield explanations are not specifically designed for CNN models. Thus, it is not surprising that these methods might not be able to give a satisfying explanation when one needs to further explore the structure of GNN models, especially challenging for a large and complicated model. To mitigate this issue, researchers in recent years start to develop explainable methods that are tailored to GNN models by taking the characteristics of graph structures into account. There are three different ways to achieve this goal: (1) interpreting GNN models by finding important subgraphs; (2) interpreting GNN models by generating new graphs while this generated graph is supposed to maintain the most informative features (e.g., nodes, node features, and edges); (3) interpreting GNN models by adding intermediate levels.

\subsection{Interpretable GNN through Subgraphs}

Interpretable GNN through subgraphs is a family of methods that use the subgraphs to add the interpretability of GNN models and it often focuses on the local features and then only yields the most important subgraph. 

Ying et al.~\cite{ying2019gnn} proposed GNNExplainer, which interprets GNN through subgraphs and is a model-agnostic approach to provide interpretable explanations for predictions of any GNN-based model. GNNExplainer identifies a compact subgraph structure and a small subset of node features that play a crucial role in GNN’s prediction. To explain a given node’s predicted label, GNNExplainer provides a local interpretation by highlighting relevant features as well as an important subgraph structure by identifying the edges that are most relevant to the prediction. This is a pioneer method in explaining GNNs. GNNExplainer  can provide explanations for any GNN that mutual-information is apt for the task, by finding both important subgraphs and important subfeatures.

After that, Zhang et al.~\cite{zhang2020relex} proposed a model-agnostic relational model explainer called RelEx, which treats the underlying model as a black-box model and learns relational explanations. The RelEx constructs explanations using two steps---learning a local differentiable approximation of the black-box model and then learning an interpretable mask over the local approximation with the use of subgraphs. It can provide flexible explanation for end users and the local approximator of GNN models is locally faithful and differentiable on the input adjacency matrix. 

In addition, Lin et al.~\cite{lin2020graph} presented a model-agnostic framework, called Graph neural networks Including SparSe inTerpretability (GISST), for interpreting important graph structure and node features, which discards the unimportant nodes and features by inducing the sparsity. The GISST deals with the input data to obtain the important subgraph and subfeatures by getting the important probability of adjacency matrix and node features matrix. Vu et al.~\cite{vu2020pgm} proposed a model-agnostic explainer called Probabilistic Graphical Model for GNNs (PGM-Explainer) by identifying crucial graph components to generate an explanation. PGM-Explainer produces a simpler interpretable Bayesian and can illustrate the dependency among explained features and provide deeper explanations for GNN’s predictions. Yuan et. al.~\cite{yuan2021explainability} proposed SubgraphX to explain GNNs by identifying important subgraphs. The information aggregation procedures in GNNs can be interpreted as interactions among different graph structures. Thus the authors used Shapley values to measure the importance of subgraphs by capturing such interactions only within the information aggregation range. Furthermore, they used Monte Carlo tree search algorithm to efficiently explore different subgraphs for a given input graph. The SubgraphX explains GNNs via identifying subgraphs explicitly.

\subsection{Interpretable GNN through Graphs Generation}

Instead of focusing on subgraphs, interpreting GNNs through graphs generation takes the whole graph structure (or global structure) into consideration. It considers the overall structure of the graph. Then a new graph is generated that contains only the structure necessary for the decision making by GNNs.

Similar to the PGM-Explainer analysing the explained features from conditional probabilities, Luo et al.~\cite{luo2020parameterized} proposed a model-agnostic method of explainable GNNs called PGExplainer. PGExplainer provides explanations for GNNs by generating a probabilistic graph. It is naturally applicable to provide model-level explanations for each instance with a global view of the GNN model and has better generalization ability. On the other hand, Yuan et al.~\cite{yuan2020xgnn} also proposed XGNN, which provides model-level explanations without preserving the local fidelity. XGNN applied reinforcement learning to generate important graph to explain the prediction which is made by GNN models. It generates graph patterns by maximizing a certain prediction of the model. Thus it can provide high-level insights and a generic understanding of how GNNs work. 



\subsection{Interpretable GNN through Intermediate Levels Injection}

Interpreting GNN through intermediate levels injection can directly encode knowledge/information as a factor graph into the model architecture. For example, the Factor Graph Neural Network (FGNN) model established by Ma et al.~\cite{ma2019incorporating} directly encodes biological knowledge such as Gene Ontology into the model architecture. Each node in the Factor Graph Neural Network model corresponds to some biological entity such as genes or Gene Ontology terms, making the model transparent and interpretable. 

In addition, Yang et al.~\cite{yang2020graph} proposed the Graph-based neural networks for Image Privacy (GIP) to infer the privacy risk of images. The GIP mainly focuses on objects in an image, and the knowledge graph is extracted from the objects in the dataset without reliance on extra knowledge. The results showed that the introduction of the knowledge graph not only makes the deep model more explainable but also makes better use of the information of objects provided by the images. Furthermore, Yu et al.~\cite{yu2021scene} proposed Scene-Graph Autonomous Driving Systems (SGADS) which used scene-graphs as intermediate representations to deal with the limitations in Autonomous Driving Systems (ADS) and the spatial and temporal attention components used in their approach improved both its performance and its explainability. Noutahi et al.~\cite{noutahi2019towards} proposed LaPool, an interpretable hierarchical graph pooling method, which uses Laplacian Pooling as an intermediate to capture the relative importance of interactions between molecular substructures. LaPool takes into account both node features and graph structure to improve molecular representation. 



\section{Evaluation Metrics for GNN Explainers}\label{sec:metric}

\begin{figure}
    \centering
    \includegraphics[width=0.95\textwidth]{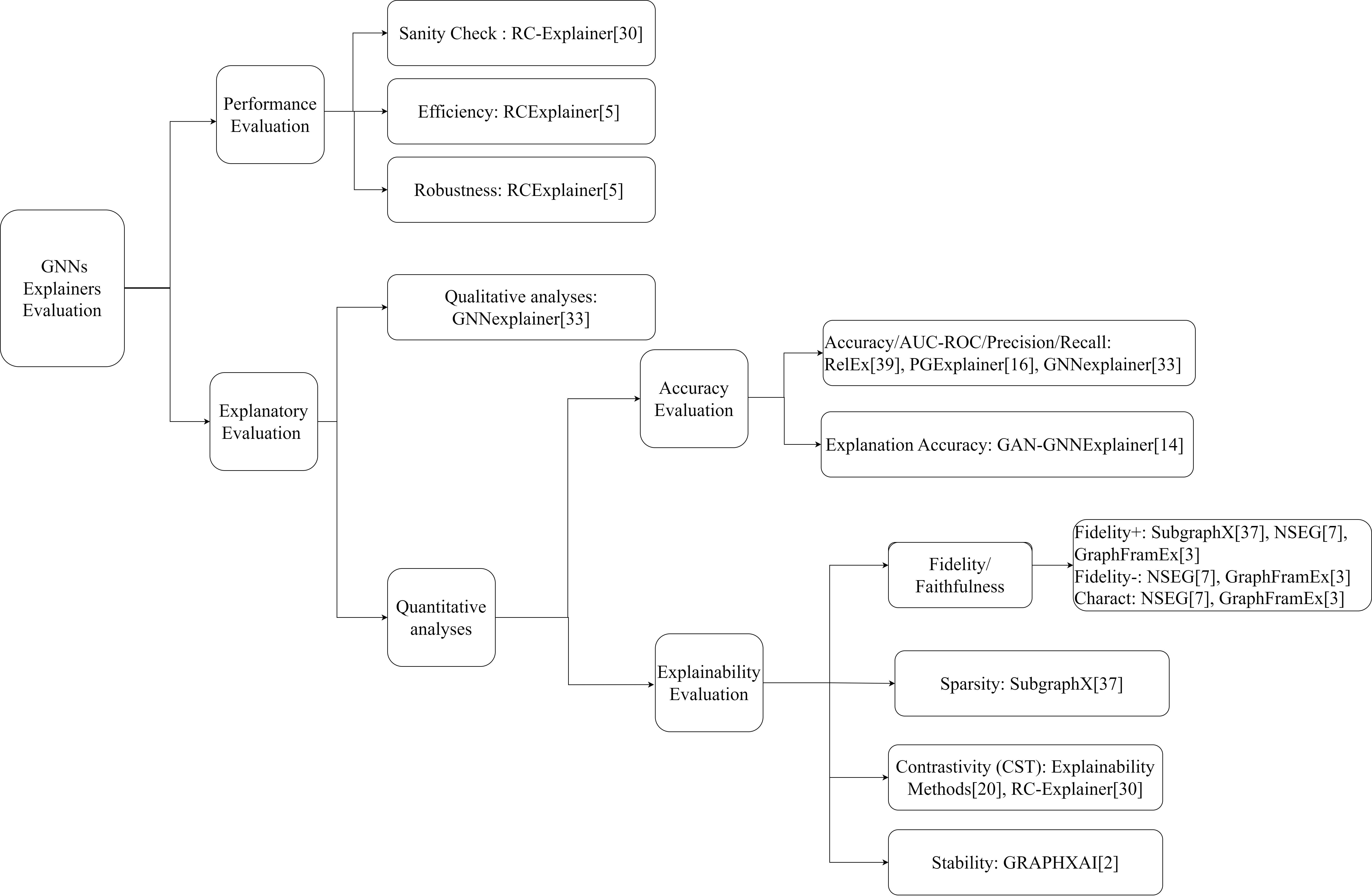}
    \caption{The taxonomy of metrics.}
    \label{fig:metrics}
\end{figure}

Since explainers are used to explain why a certain decision has been made instead of depicting the whole black-box, there is uncertainty about the fidelity of the explainer itself. Therefore, it is crucial to use the right metrics to evaluate the correctness and completeness of the interpretability techniques. Recently, GraphFramEx~\cite{GraphFramEx_2022} and GRAPHXAI~\cite{GRAPHXAI_2022} have focused on  defining the explainability metrics to evaluate the fidelity of GNNs explanations. Further, some evaluation metrics for XAI~\cite{2021Evaluating} are also available to be applied to GNN explainers. This section provides a short review of the prevalent evaluation metrics for GNNs explanations. Generally, we evaluate a GNNs explainer from two aspects: performance and explanatory capability. In specific, explanatory capability can be evaluated from qualitative analyses and quantitative analyses, including accuracy evaluation and explainability evaluation. The taxonomy of metrics can be found in figure~\ref{fig:metrics}. 

\subsection{Performance Evaluation}

\paragraph{Efficiency.} An efficient graph explanation algorithm should be able to provide explanations for a large number of decisions made by a machine learning model quickly and with minimal computational resources. This is particularly important in scenarios where real-time decision-making is required or where the volume of data is extremely large. In addition to being time and resource-efficient, an efficient graph explanation algorithm should also produce explanations that are accurate, interpretable, and fair. Achieving a balance between efficiency and accuracy/fairness is an active research area in the field of graph explanation. In the paper~\cite{RCExplainer}, authors evaluate efficiency by comparing the average computation time taken for inference on unseen graph samples.

\paragraph{Sanity Check (SC).} Explainers should provide explanations for GNNs by finding both important subgraphs and important features that play a crucial role in the prediction of GNN. Thus, good GNNs explainers should be sensitive to the target GNN model changes. A sanity check~\cite{Reinforced_Causal_Explainer} is one way to evaluate the sensitivity capability of GNNs explainers. In specific, SC is to compare the attribution scores on the trained GNN $f$ with that on an untrained GNN $\hat{f}$ with randomly-initialized parameters. Similar attributions infer the explainer is insensitive to the properties of the model, thus failing to pass the check. Thus, we desire a ~\emph{lower} $SC$ ($\downarrow$). The definition of SC is shown in~\cref{eq:sc}.

\begin{equation}\label{eq:sc}
\mathrm{SC}=\mathbb{E}_{\mathcal{G} \sim \mathbb{G}}[|\rho(\Phi(\mathcal{G}, f(\mathcal{G})), \Phi(\mathcal{G}, \tilde{f}(\mathcal{G})))|]
\end{equation}

\paragraph{Robustness.} It means the explanations of interpretation methods resist attacks such as input corruption/perturbation, adversarial attack and model manipulation. A robust interpretation method can provide similar explanations despite the presence of such attacks~\cite{luo2020parameterized,zhang2020relex}. Authors~\cite{RCExplainer} computer robustness by  quantifying how much an explanation changes after adding noise to the input graph.

\subsection{Explanatory Evaluation: Qualitative analyses}

\paragraph{Qualitative analyses.} Qualitative analyses are an important aspect of explainability in Graph Neural Networks (GNNs). These analyses involve examining the internal workings of a GNN to gain insight into how it makes decisions. This can be accomplished through techniques such as visualization, feature importance analysis, and interpretation of node and edge embeddings. By conducting qualitative analyses, researchers and practitioners can better understand the factors that contribute to GNN decision-making, identify potential biases or errors, and improve the overall transparency and interpretability of the model. Ultimately, qualitative analyses are crucial for ensuring that GNNs are trustworthy and can be used effectively in real-world applications. Qualitative analyses have been widely used in recent research, such as GNNExplainer~\cite{ying2019gnn}, PGExplainer~\cite{luo2020parameterized}, GAN-GNNExplainer~\cite{GANExplainer}, etc.

\subsection{Explanatory Evaluation: Quantitative analyses}

\subsubsection{Accuracy Evaluation}
Accuracy evaluation refers to the process of assessing the correctness and fidelity of the explanations generated by an algorithm or model. Accurate explanations are essential for building trust in the machine learning model's decision-making process and for ensuring fairness and transparency. Therefore, accuracy evaluation is a crucial step in developing and evaluating graph explanation algorithms.


\paragraph{Accuracy (ACC).} ACC is the proportion of explanations that are "correct". There are two definitions to measure the \emph{accuracy} of explainable methods. First, one can use the percentage of the identified important features (e.g., nodes, node features, and edges) to the true important truth~\cite{luo2020parameterized,ying2019gnn,zhang2020relex} (see~\cref{eq:acc_1}):
\begin{equation}\label{eq:acc_1}
Accuracy = \frac{1}{N}\sum^{N}_{i=1} \frac{|s_i|}{|S_i|_{gt}}
\end{equation}
where $|S_i|_{gt}$ represents the truth important number of features; while $|s_i|$ is the important features identified by the explainable methods; $N$ is the total number of samples. While this approach is simple and intuitive, however, it requires the ground-truth explanations of datasets, which is often hard to obtain in the real world. The other one is explanation accuracy.


\paragraph{Explanation Accuracy.} This is derived from the perspective of model predictions and measures the prediction accuracy~\cite{GANExplainer}. They use the predictions of the target GNN for the explanations to calculate the accuracy of the explanation. The accuracy of the explanation can be defined as~\cref{eq:acc_exp}:

\begin{equation}\label{eq:acc_exp}
ACC_{exp}=\frac{|f(\mathbf{G})=f(\mathbf{G}^s)|}{|T|}
\end{equation}

where $f$ is the pre-trained target GNN, $\mathbf{G}$ is the original graph we want to explain, and $\mathbf{G}^s$ is its corresponding explanation (e.g., the important subgraph), $|f(\mathbf{G})=f(\mathbf{G}^s)|$ is the corrected classified number which means $f(\mathbf{G})=f(\mathbf{G}^s)$, $|T|$ is the total number of the test set.




\subsubsection{Explainability Evaluation} 
\paragraph{Fidelity.} It measures whether the explanations are faithfully important to the model’s predictions. The $Fidelity^+$~\cite{yuan2020explainability,yuan2021explainability} metric indicates the difference in predicted probability between the original predictions and the new prediction after removing important input features. In contrast, the metric $Fidelity^-$~\cite{yuan2020explainability} represents prediction changes by keeping important input features and removing unimportant structures. 

\begin{equation}\label{eq:fidelity+}
Fidelity^+ = \frac{1}{N}\sum^{N}_{i=1} (f(\bm{G}_i)_{y_i}-f(\bm{G}_i^{1-m_i})_{y_i})
\end{equation}

\begin{equation}\label{eq:fidelity-}
Fidelity^- = \frac{1}{N}\sum^{N}_{i=1} (f(\bm{G}_i)_{y_i}-f(\bm{G}_i^{m_i})_{y_i})
\end{equation}

Where $N$ is the total number of samples, and $y_i$ is the class label. $f(\bm{G}_i)_{y_i}$ and $f(\bm{G}_i^{1-m_i})_{y_i}$ are the prediction probabilities of $y_i$ when using the original graph $\bm{G}_i$ and the occluded graph $\bm{G}_i^{1-m_i}$, which is gained by occluding important features found by explainers from the original graph. Thus, a~\emph{higher} $Fidelity^+$ ($\uparrow$) is desired. $f(\bm{G}_i^{m_i})_{y_i}$ is the prediction probabilities of $y_i$ when using the explanation graph $\bm{G}_i^{m_i}$, which is obtained by important structures found by explainable methods. Thus a~\emph{lower} $Fidelity^-$ ($\downarrow$) is desired. Specifically, $Fidelity^+$ and $Fidelity^-$ are used to quantify the necessity and sufficiency of the explanations, respectively. The higher $Fidelity^+$, the more necessary the explanation. On the contrary, the lower $Fidelity^-$, the more sufficient the explanation.

\paragraph{Characterization Score.} The characterization score~\cite{GraphFramEx_2022,NSEG} is a global evaluation metric that attempts to balance the sufficiency and necessity requirements. This approach is analogous to combining precision and recall in the Micro-F1 metric. The characterization score is the weighted harmonic mean of Fidelity+ and Fidelity- as defined below:

\begin{equation}\label{eq:charact}
Charact = \frac{2 \times Fidelity^+\times(1-Fidelity^-)}{Fidelity^+ +(1-Fidelity^-)}
\end{equation}

\paragraph{Sparsity.} It measures the fraction of features selected as important by explanation methods~\cite{pope2019explainability,yuan2021explainability}, which is defined in~\cref{eq:sparsity}: 

\begin{equation}\label{eq:sparsity}
Sparsity = \frac{1}{N}\sum^{N}_{i=1} (1-\frac{|s_i|}{|S_i|_{total}})
\end{equation}
where the $|S_i|_{total}$ represents the total number of features (e.g., nodes, nodes features, or edges) in the original graph model; while $|s_i|$ is the size of important features/nodes found by the explainable methods and it is a subset of $|S_i|$; $N$ is the total number of samples. Note that higher sparsity values indicate that explanations are sparser and likely to capture only the most essential input information. Hence, a~\emph{higher} $Sparsity$ ($\uparrow$) is desired.


\paragraph{Contrastivity (CST).} CST means the ratio of the Hamming distance between binarized heat-maps for positive and negative classes~\cite{pope2019explainability}. The underlying idea behind contrastivity is that the highlighted features by an explanation method should vary across classes. ~\cite{pope2019explainability} and ~\cite{Reinforced_Causal_Explainer} defined and used CST to evaluate the explainability of their methods. One can define fidelity as shown in~\cref{eq:cst}. And a~\emph{lower} $CST^-$ ($\downarrow$) is desired.

\begin{equation}\label{eq:cst}
\mathrm{CST}=\mathbb{E}_{\mathcal{G} \sim \mathbb{G}} \mathbb{E}_{s \neq \hat{y}}[\rho(\Phi(\mathcal{G}, s), \Phi(\mathcal{G}, \hat{y})) \mid]
\end{equation}

\paragraph{Stability.} Graph explanation stability refers to the ability of a Graph Neural Network (GNN) to produce consistent explanations even when the input graph is slightly altered or perturbed. This is important for ensuring the reliability and interpretability of the model's decisions. In~\cite{GRAPHXAI_2022}, authors measure graph explanation stability by computer the instability degree. They calculate the instability as~\cref{eq:stability}. 
\begin{equation}\label{eq:stability}
\operatorname{GES}\left(\mathbf{M}_{\mathcal{S}_{u^{\prime}}}^p, \mathbf{M}_{\mathcal{S}_{u}^p}\right)=\max D\left(\mathbf{M}_{\mathcal{S}_{u^{\prime}}^p}, \mathbf{M}_{\mathcal{S}_{u^{\prime}}^p}\right), \quad \forall \mathcal{S}_{u^{\prime}} \in \beta\left(\mathcal{S}_u\right)
\end{equation}

In here, $\mathcal{S}_u$ is the subgraph of node $u$, and the $\mathcal{S}_{u^{\prime}}$ is the subgraph of perturbed node $u^{\prime}$; $\max D$ represents the cosine distance metric, $\mathbf{M}_{\mathcal{S}_{u}^p}$ and $\mathbf{M}_{\mathcal{S}_{u^{\prime}}^p}$ are the predicted explanation masks for $\mathcal{S}_u$ and $\mathcal{S}_{u^{\prime}}$; and $\beta$ represents a $\delta$-radius ball around $\mathcal{S}_u$ for which the model behavior is same. 

\paragraph{Fairness.} It is the concept that explanations provided by machine learning models should be accurate and fair, and should not perpetuate or amplify existing biases. It promotes transparency, accountability, and fairness in decision-making processes. In the paper~\cite{GRAPHXAI_2022}, authors propose Graph Explanation Counterfactual fairness mismatch (GECF) and Graph Explanation Group Fairness mismatch (GEGF) to evaluate the explanations on the respective datasets. To measure counterfactual fairness, they verify if the explanations corresponding to $\mathcal{S}_{u}$ and its counterfactual counterpart are similar if the underlying model predictions are similar. They calculate counterfactual fairness mismatch as:
\begin{equation}
\operatorname{GECF}\left(\mathbf{M}^p, \mathbf{M}_s^p\right)=D\left(\mathbf{M}^p, \mathbf{M}_s^p\right)
\end{equation}

where $\mathbf{M}^p$ and $\mathbf{M}_s^p$ are the predicted explanation mask for $\mathcal{S}_{u}$, and for the counterfactual counterpart of $\mathcal{S}_{u}$. It should be noted that they anticipate a decrease in the GECF score for graphs that have ground-truth explanations that exhibit weak forms of unfairness. This is because the explanations for both the original and counterfactual graphs are likely to be similar. In contrast, for graphs with ground-truth explanations that exhibit strong forms of unfairness, we expect to observe an increase in the GECF score. This is because modifying the protected attribute is likely to result in changes to the explanations provided by the model.

 They measure group fairness mismatch as follows:
\begin{equation}
\operatorname{GEGF}\left(\hat{\mathbf{y}}_{\mathcal{K}}, \hat{\mathbf{y}}_{\mathcal{K}}^{\mathbf{E}_u}\right)=\left|\operatorname{SP}\left(\hat{\mathbf{y}}_{\mathcal{K}}\right)-\operatorname{SP}\left(\hat{\mathbf{y}}_{\mathcal{K}}^{\mathbf{E}_u}\right)\right|
\end{equation}

Where $\hat{\mathbf{y}}_{\mathcal{K}}$ and $\hat{\mathbf{y}}_{\mathcal{K}}^{\mathrm{E}_u}$ are  predictions for a set of K graphs using the original and the essential features identified by an explanation, respectively. And $\operatorname{SP}$ is the statistical parity. The higher values of GEGF indicate that the explanation is not preserving group fairness.

There are various evaluation metrics, and each one has its respective emphasis and reflects different aspects of an explainable model. One should therefore use a combination of multiple metrics to attain reasonable and practical explainable systems. However, as mentioned above, it is also important that one should take the characteristics of datasets and explainable methods into account in order to choose suitable evaluation metrics.

\section{Discussion}\label{sec:dis}
This survey focuses on providing a clear taxonomy of explainable GNNs. After analyzing the literature on explainable GNNs, we summarized the problems as shown below.

\begin{itemize} 
\item \textit{How to explain graph neural networks?} There are two major perspectives.
\begin{itemize}
    \item GNNs can be treated as a black-box and find an independent way to explain the links between input and output, such as GraphLIME or RelEx.
    \item Another way tries to explain the details of the GNN by leveraging the information of nodes and edges in itself.
\end{itemize}

\item \textit{How to extend XAI methods to graph neural networks?}
There are some studies using the XAI methods to explain GNNs (see Section~\ref{subsec:XAI-GNNS}). The XAI methods including Saliency Maps, LRP, LIME, Guided BP, Grad-CAM, and etc., get a competitive performance for XAI and they can be extended to explain GNNs. However, those methods are not specifically designed for GNNs and require knowledge on the internal parameters of the model. 

\item \textit{How to find the most important subgraph structure that influences the predictions of graph neural networks?} As we mentioned in  section 3.2, there are several methods to explain GNNs by focusing subgraph structures. For example, the GNNExplainer identifies a compact subgraph structure and a small subset of node features that may play a crucial role in GNN’s prediction. Furthermore, the PGMExplainer and GISST generate explanations by yielding an important subgraph and node feature subset related to any graph-based task. However, these methods only focus on subgraph structures which are local information and fail to consider any global features.

\item \textit{How to explain graph neural networks from a global perspective?} Instead of the segmented information obtained through local graph structure, the global structure can often provide more interesting and complete information. For instance, the PGExplainer focuses on the explanation of the complete graph structures and provides a global understanding of predictions made by GNNs. It can explain predictions of GNNs on a set of instances collectively and easily generalize the learned explainer model to other instances. 
\end{itemize}

\section{Conclusion and Future Work}\label{sec:con}
By analyzing recent studies of XGNNs in details, we observe that we can not only increase the interpretability of GNNs with direct use of XAI methods, but also can create novel XGNNs methods without involving current XAI approaches to improve GNN models interpretability or transparency. We also find that studies in explainable GNNs are still in their early stages. We, therefore, present the following future research directions.

The ultimate goal of XGNNs is to provide clues of how a GNN model makes its decisions to human, so that human can trust its prediction or not. Human hence play a vita role in this explainable loop. It is therefore an interesting research direction to incorporate the human experience (knowledge and feedback) into this procedure to achieve better explainable algorithms or models.

Although there are several work that leverage the existing XAI methods to achieve the explainability of GNNs, there is no unify rules to guide this procedure. It is worth developing such rules so that more existing XAI methods can be contributed to the filed of XGNNs. Furthermore, each XAI method has its pros and cons, it is therefore interesting to explore different combinations of different XAI methods (e.g., in an ensemble manner) to obtain better explainability.


\bibliographystyle{splncs04}

\bibliography{survey.bib}

\end{document}